\documentclass{article}
\usepackage{microtype}
\usepackage{graphicx}
\usepackage{subcaption}
\usepackage{booktabs} 
\usepackage{amsmath,amsfonts,amsthm,bm, amssymb}
\usepackage{hyperref}
\usepackage[capitalise]{cleveref}
\usepackage{enumitem}
\usepackage{booktabs}
\usepackage[T1]{fontenc}
\usepackage{xfrac}
\theoremstyle{definition}
\DeclareMathOperator*{\argmax}{arg\,max}
\newtheorem{definition}{Definition}

\newcommand{\D}{D(\textbf{x}; \bm{\theta}^{(D)})}
\newcommand{\G}{G(\textbf{z}; \bm{\theta}^{(G)})}
\usepackage[accepted]{icml2018}

\DeclareMathOperator{\RLT}{RLT}
\DeclareMathOperator{\GeomScore}{GeomScore}
\DeclareMathOperator{\mRLT}{MRLT}
\icmltitlerunning{Geometry Score: A Method For Comparing GANs}
\begin{document}

\twocolumn[
\icmltitle{Geometry Score: A Method For Comparing Generative Adversarial Networks}
\icmlsetsymbol{equal}{*}
\begin{icmlauthorlist}
\icmlauthor{Valentin Khrulkov}{skol}
\icmlauthor{Ivan Oseledets}{skol,ivm}
\end{icmlauthorlist}
\icmlaffiliation{ivm}{Institute of Numerical Mathematics RAS}
\icmlaffiliation{skol}{Skolkovo Institute of Science and Technology}
\icmlcorrespondingauthor{Valentin Khrulkov}{khrulkov.v@gmail.com}
\icmlkeywords{Machine Learning, GANs, Topology, TDA}
\vskip 0.3in
]
\printAffiliationsAndNotice{}
\begin{abstract}
One of the biggest challenges in the research of generative adversarial networks (GANs) is assessing the quality of generated samples and detecting various levels of mode collapse. In this work, we construct a novel measure of performance of a GAN by comparing geometrical properties of the underlying data manifold and the generated one, which provides both qualitative and quantitative means for evaluation. Our algorithm can be applied to datasets of an arbitrary nature and is not limited to visual data. We test the obtained metric on various real--life models and datasets and demonstrate that our method provides new insights into properties of GANs.
\end{abstract}
\section{Introduction}
Generative adversarial networks (GANs) \cite{goodfellow_gan} are a class of methods for training generative models, which have been recently shown to be very successful in producing image samples of excellent quality. They have been applied in numerous areas \cite{radford2015unsupervised, improved_techniques, ho-ermon-gen-adv}. Briefly, this framework can be described as follows. We attempt to mimic a given target distribution $p_{\text{data}} (\textbf{x})$ by constructing two networks $\G$ and $\D$ called the generator and the discriminator. The generator learns to sample from the target distribution by transforming a random input vector $\textbf{z}$ to a vector $\textbf{x} = \G$, and the discriminator learns to distinguish the model distribution $p_{\text{model}}(\textbf{x})$ from $p_{\text{data}} (\textbf{x})$. The training procedure for GANs is typically based on applying gradient descent in turn to the discriminator and the generator in order to minimize a loss function. Finding a good loss function is a topic of ongoing research, and several options were proposed in \cite{mao2016least, arjovsky2017wasserstein}.

One of the main challenges \cite{lucic2017gans, barratt2018note} in the GANs framework is estimating the quality of the generated samples. In traditional GAN models, the discriminator loss cannot be used as a metric and does not necessarily decrease during training. In more involved architectures such as WGAN \cite{arjovsky2017wasserstein} the discriminator (critic) loss is argued to be in correlation with the image quality, however, using this loss as a measure of quality is nontrivial. Training GANs is known to be difficult in general and presents such issues as \emph{mode collapse} when $p_{\text{model}}(\textbf{x})$ fails to capture a multimodal nature of $p_{\text{data}}(\textbf{x})$ and in extreme cases all the generated samples might be identical. Several techniques to improve the training procedure were proposed in \cite{improved_techniques, gulrajani2017improved}.

\begin{figure}[htb!]
\centering
\includegraphics[width=0.9\linewidth]{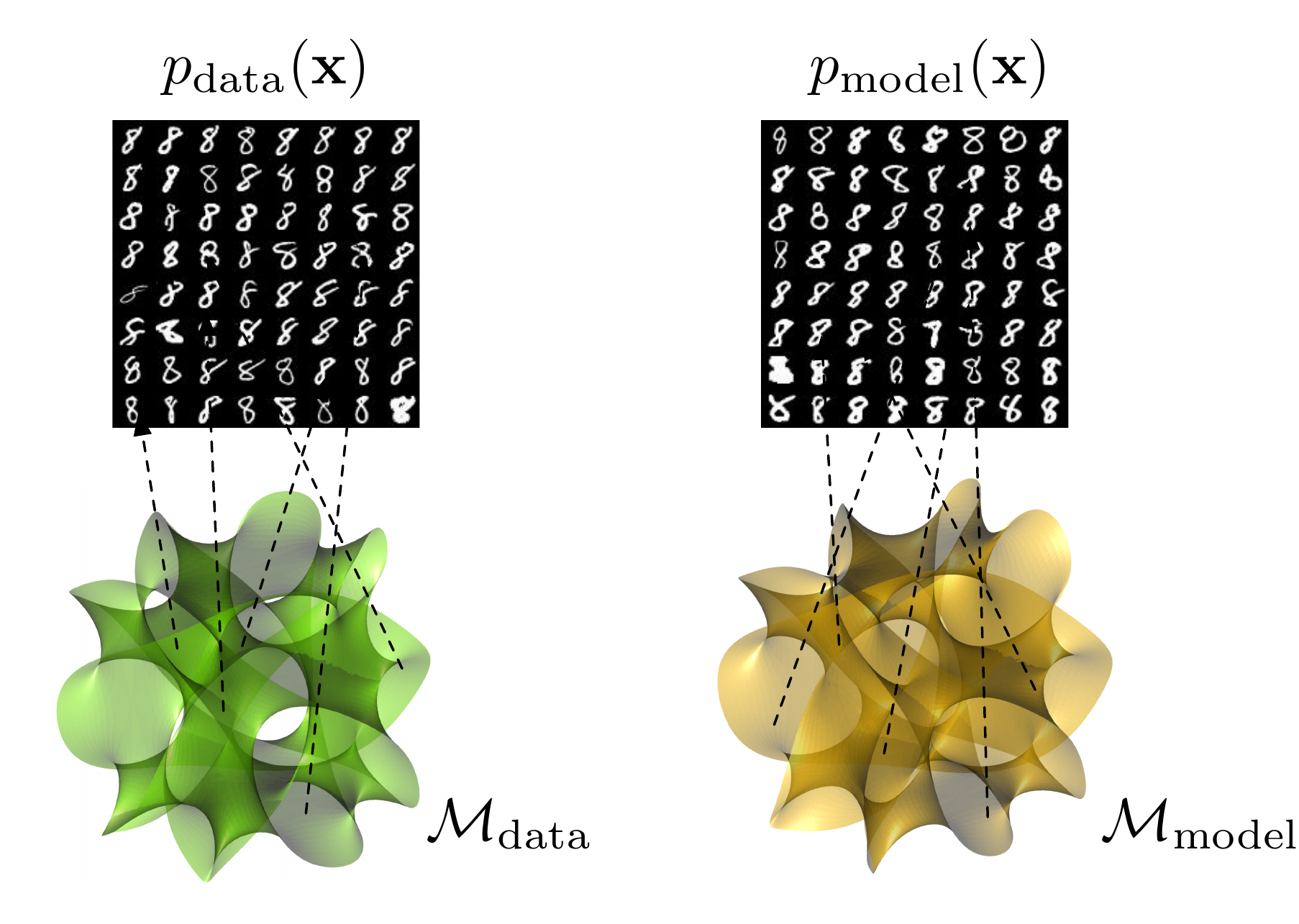}
\caption{The Manifold Hypothesis suggests that in the case of natural images the data is supported on a low dimensional \emph{data manifold} $\mathcal{M}_{\text{data}}$. Similarly, GANs sample images from an immersed manifold $\mathcal{M}_{\text{model}}$. By comparing topological properties of the manifolds $\mathcal{M}_{\text{data}}$ and $\mathcal{M}_{\text{model}}$ we can get insight in how strongly GAN captured intricacies in the data distribution $p_{\text{data}}(\textbf{x})$, and quantitatively estimate the difference.}
\label{fig:data-manifolds}
\end{figure}

In this work, we attack the problem of estimating the quality and diversity of the generated images by using the machinery of topology. The well-known Manifold Hypothesis \cite{goodfellow2016deep} states that in many cases such as the case of natural images the support of the distribution $p_{\text{data}}(\textbf{x})$ is concentrated on a low dimensional manifold $\mathcal{M}_{\text{data}}$ in a Euclidean space. This manifold is assumed to have a very complex non-linear structure and is hard to define explicitly. It can be argued that interesting features and patterns of the images from $p_{\text{data}}(\textbf{x})$ can be analyzed in terms of topological properties of $\mathcal{M}_{\text{data}}$, namely in terms of loops and higher dimensional \emph{holes} in $\mathcal{M}_{\text{data}}$. Similarly, we can assume that $p_{\text{model}}(\textbf{x})$ is supported on a manifold $\mathcal{M}_{\text{model}}$ (under mild conditions on the architecture of the generator this statement can be made precise \cite{shao2017riemannian}), and for sufficiently good GANs this manifold can be argued to be quite similar to $\mathcal{M}_{\text{data}}$ (see 
\cref{fig:data-manifolds}). This intuitive claim will be later supported by numerical experiments. Based on this hypothesis we develop an approach which allows for comparing the topology of the underlying manifolds for two point clouds in a stochastic manner providing us with a visual way to detect mode collapse and a score which allows for comparing the quality of various trained models. Informally, since the task of computing the precise topological properties of the underlying manifolds based only on samples is ill-posed by nature, we estimate them using a certain probability distribution (see \cref{sec:stochastic}).

We test our approach on several real--life datasets and popular GAN models (DCGAN, WGAN, WGAN-GP) and show that the obtained results agree well with the intuition and allow for comparison of various models (see \cref{sec:experiments}).
\section{Main idea \label{sec:idea}}
\begin{figure}[htb!]
\centering
\includegraphics[width=0.8\linewidth]{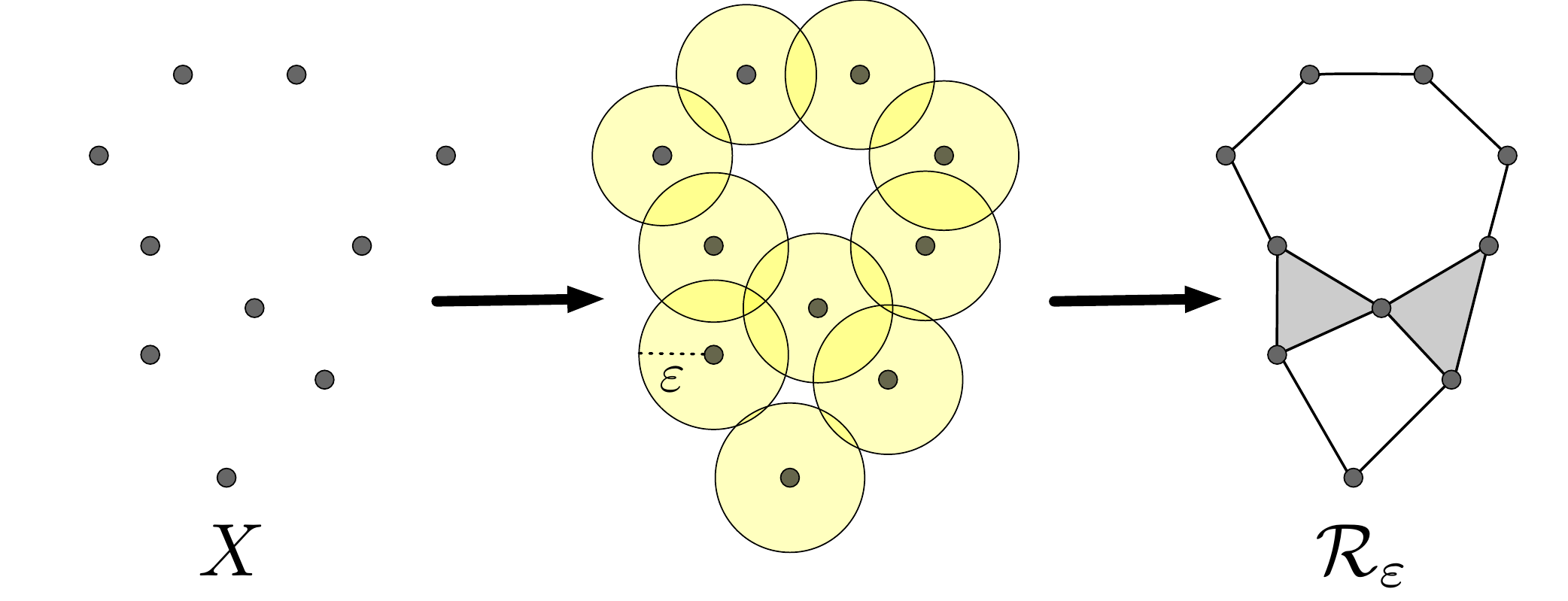}
 \caption{A simplicial complex constructed on a sample $X$. First, we fix the proximity parameter $\varepsilon$. Then we take balls of the radius $\varepsilon$ centered at each point, and if for some subset of $X$ of size $k+1$ all the pairwise intersections of the corresponding balls are non-empty, we add the $k$-dimensional simplex spanning this subset to the simplicial complex $\mathcal{R}_{\varepsilon}$.}
  \label{fig:rips_compl}
\end{figure}

Let us briefly discuss our approach before dwelling into technical details. As described in the introduction we would like to compare topological properties of $\mathcal{M}_{\text{data}}$ and $\mathcal{M}_{\text{model}}$ in some way. This task is complicated by the fact that we do not have access to the manifolds themselves but merely to samples from them. A natural approach in this case is to approximate these manifolds using some simpler spaces in such a way that topological properties of these spaces resemble those of $\mathcal{M}_{\text{data}}$ and $\mathcal{M}_{\text{model}}$. 
\begin{figure}[htb!]
\centering
	\includegraphics[width=0.5\linewidth]{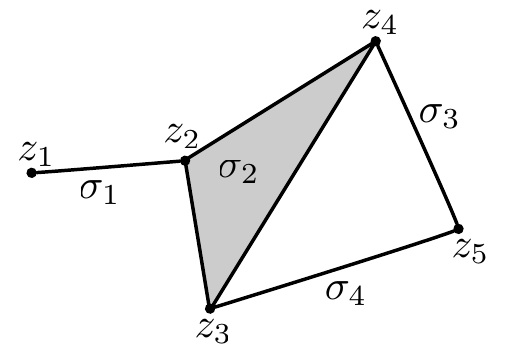}
 \caption{Simplicial complex. Topological space $X$ is constructed from several edges ($\sigma_1, \sigma_3, \sigma_4$) and a two dimensional face $\sigma_2$.}
  \label{fig:simp_compl}
\end{figure}
The main example of such spaces are \textbf{simplicial complexes} (\cref{fig:simp_compl}), which are build from intervals, triangles and other higher dimensional simplices. In order to reconstruct the underlying manifold using a simplicial complex several methods exist. In all such approaches proximity information of the data is used, such as pairwise distances between samples. Typically one chooses some threshold parameter $\varepsilon$ and based on the value of this parameter one decides 
which simplices are added into the approximation (see \cref{fig:rips_compl}). 
\begin{figure}[htb!]
\centering
\includegraphics[width=0.8\linewidth]{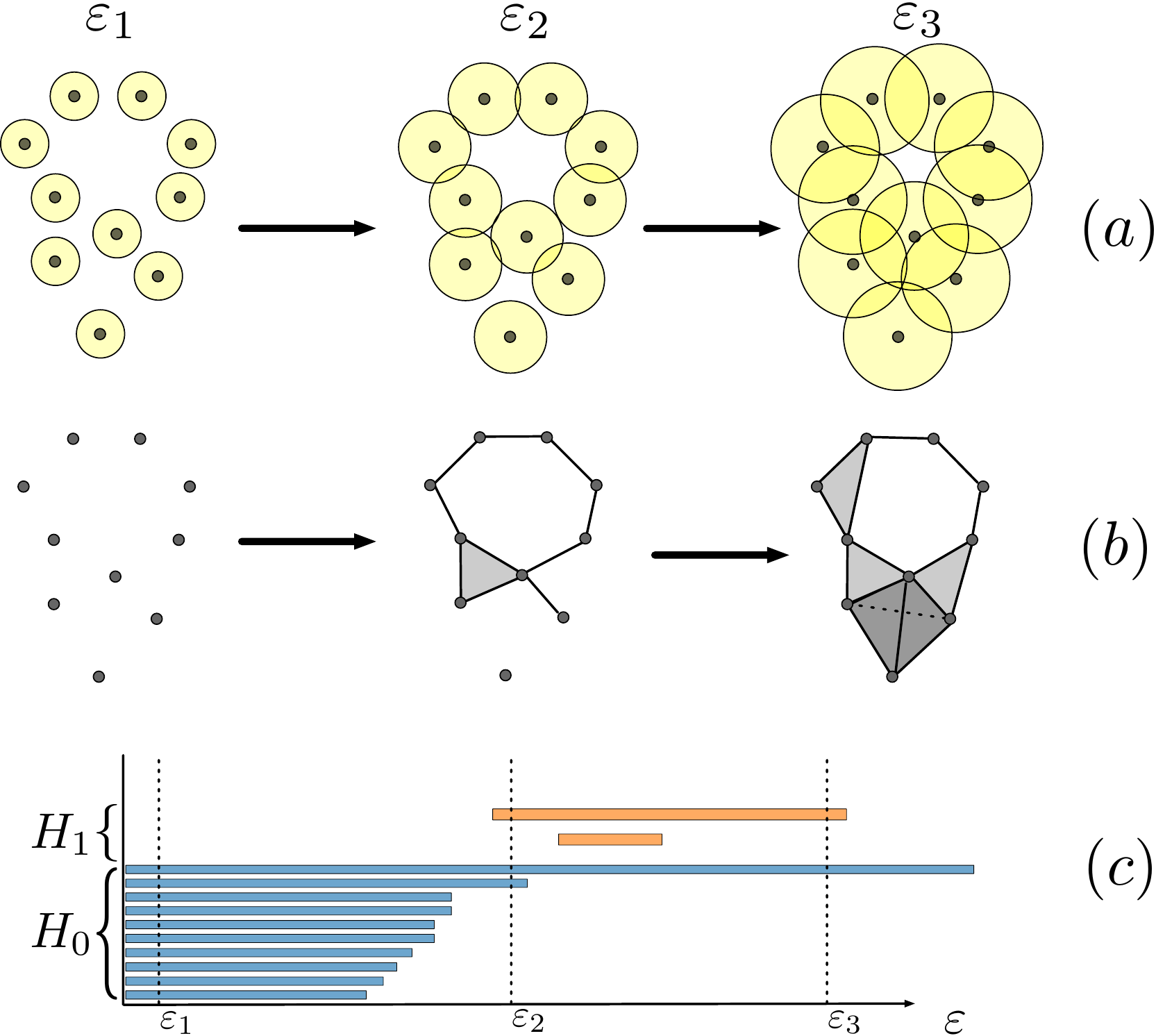}
 \caption{Using different values of the proximity parameter $\varepsilon$ we obtain different simplicial complexes (a). For $\varepsilon = \varepsilon_1$ the balls do not intersect and there are just $10$ isolated components (b, [left]). For $\varepsilon = \varepsilon_2$ several components have merged and one \emph{loop} appeared (b, [middle]). The filled triangle corresponding to the triple pairwise intersection is topologically trivial and does not affect the topology (and similarly darker tetrahedron on the right). For $\varepsilon = \varepsilon_3$ all the components merged into one and the same hole still exists (b, [right]). In the interval $[\varepsilon_2, \varepsilon_3]$ one smaller hole as on \cref{fig:rips_compl} appeared and quickly disappeared. This information can be conveniently summarized in the \emph{persistence barcode} (c). The number of connected components (holes) in the simplicial complex for some value $\varepsilon_0$ is given by the number of intervals in $H_0$ ($H_1$) intersecting the vertical line $\varepsilon=\varepsilon_0$.}
  \label{fig:sc-family}
\end{figure}
However a single value $\varepsilon$ is not enough --- for very small values the reconstructed space will be just a disjoint union of points and for very large $\varepsilon$ it will be a single connected blob, while the correct approximation is somewhere in between. This issue is resolved by considering a \emph{family} (\cref{fig:sc-family}, a) of simplicial complexes, parametrized by the (`persistence') parameter $\varepsilon$. It is also convenient to refer to the parameter $\varepsilon$ as \emph{time}, with the idea that we gradually throw more simplices into our simplicial complex as time goes by. For each value of $\varepsilon$ we can compute topological properties of the corresponding simplicial complex, namely \emph{homology} which encodes the number of holes of various dimensions in a space. Controlling the value of $\varepsilon$ allows us to decide holes of which size are meaningful and should not be discarded as a noise. For simplicial complex presented on \cref{fig:rips_compl} there are two one-dimensional holes, and for slightly bigger value of $\varepsilon$ the lower hole disappeared (\cref{fig:sc-family}, b), while the top one remained intact, which suggests that the top hole is more important topological feature. Information about how homology is changing with respect to $\varepsilon$ can be conveniently encoded in the so-called \emph{persistence barcodes} \cite{ghrist2008barcodes,zomorodian2005computing}. An example of such barcode is given on (\cref{fig:sc-family}, c). In general, to find the rank of $k$-homology (delivering the number of $k$-dimensional holes) at some fixed value $\varepsilon_0$ one has to count intersections of the vertical line $\varepsilon = \varepsilon_0$ with the intervals at the desired block $H_k$. 

These barcodes provide a way to compare topological properties of the underlying manifolds. In principle, we could obtain a metric of similarity of two datasets by comparing the barcodes of the simplicial complexes constructed based on each dataset (as described on \cref{fig:rips_compl}), but there are disadvantages of this approach, such as a huge number of simplices for large datasets. Moreover, in order to extract interesting topological properties from such large simplicial complexes various tricks are required \cite{ghrist2008barcodes}. To remedy these issues we can note that we are in fact interested in \emph{topological} approximations rather than \emph{geometrical}. The difference is that to obtain a correct estimate of the topological properties much smaller number of simplices is often sufficient, e.g., for any number of points sampled from a circle the correct answer could be obtained by taking just three points (thus obtaining a triangle which is topologically equivalent to the circle). Based on these ideas the so-called \textbf{witness complex} is introduced \cite{de2004topological}, which provides a topological approximation with a small number of simplices. In order to achieve this a small subset of \emph{landmark} points is chosen and a simplicial complex is constructed using these points as vertices (while also taking into account the proximity information about all the remaining points called \emph{witnesses}). 

To construct a numerical measure which could be compared across datasets we would like to estimate the correct values of homology. Comparing the computed barcodes is a challenging task since they are non-trivial mathematical objects (though some metrics exist they are hard to compute). We take the simpler route and to extract meaningful topological data from the barcode we propose computing \textbf{Relative Living Times} (RLT) of each number of holes that was observed. They are defined as the ratio of the total time when this number was present and of the value $\varepsilon_{\max}$ when points connect into a single blob. These relative living times could be interpreted as a \emph{confidence} in our approximation --- if say for $50 \%$ of all period of topological activity we have observed that there is at least $1$ one-dimensional hole (as on \cref{fig:sc-family}), then it is probably an accurate estimation of topology of the underlying space.

Choosing the correct landmarks is a nontrivial task. We follow the discussion in \cite{de2004topological} which advises doing it randomly. To account for this randomness, we compute the RLT stochastically by repeating the experiment a large number of times. By averaging the obtained RLT we compute the \textbf{Mean Relative Living Times} (MRLT).
\begin{figure}[t!]
\centering
\includegraphics[width=1.0\linewidth]{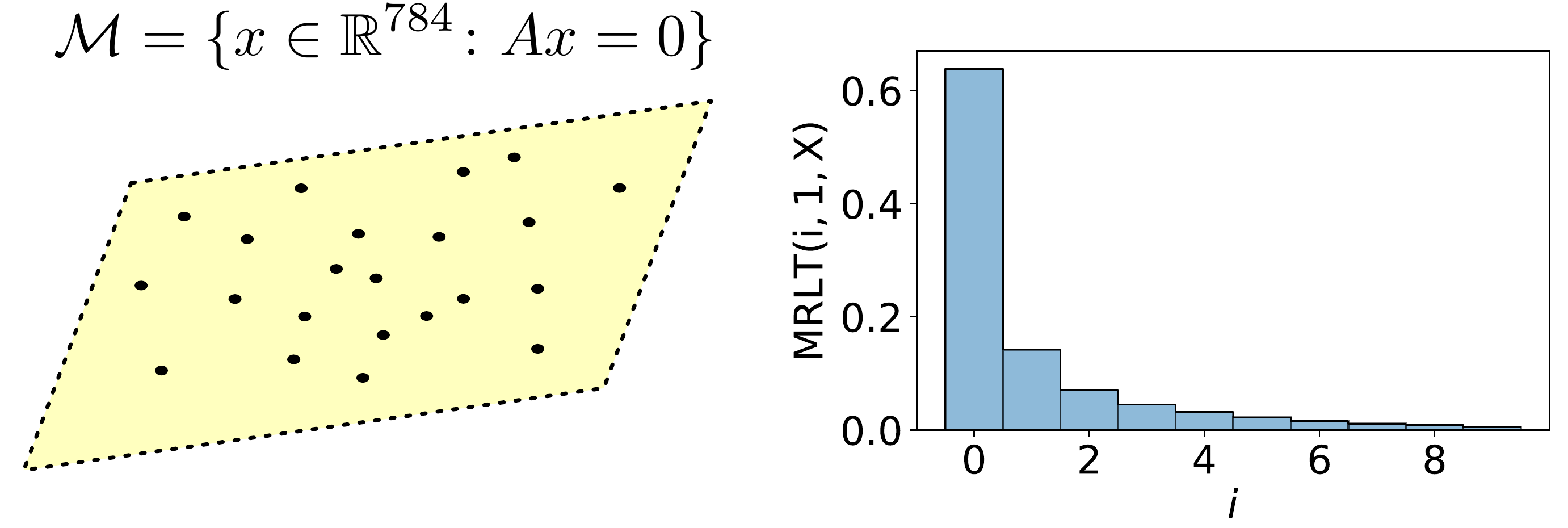}
  \caption{Estimation of the topology of a dataset sampled from the $32$-dimensional hyperplane in $784$-dimensional space. With high confidence, we can say that there are no $1$-dimensional holes. For details see \cref{sec:stochastic}.}
\label{fig:plane-topology}
\end{figure}
By construction, they add up to $1$ and employing Bayesian point of view we can interpret them as a probability distribution reflecting our confidence about the correct number of holes on \emph{average}. An example of such distribution is given on \cref{fig:plane-topology}, where we run our method for a simple planar dataset (in a high dimensional space). To quantitatively evaluate the topological difference between two datasets we propose computing the $L_2$--error between these distributions. Note that in practice (when activation functions such as ReLU are used) the resulting space $\mathcal{M}_{\text{model}}$ may fail to be a manifold in precise mathematical sense, however, the analysis is still applicable since it deals with arbitrary topological spaces.
Now let us introduce all the technical details.
\section{Homology to the rescue}
In this section we briefly discuss the important concepts of simplicial complexes and homology. For thorough introduction we refer the reader to the classical texts such as \cite{hatcher2002algebraic,may1999concise}.
\paragraph{Simplicial complexes} 
Simplicial complex is a classical concept widely used in topology. Formally it is defined as follows. 
\begin{definition}
A simplicial complex $\mathcal{S}$ (more precisely an \emph{abstract} simplicial complex) is specified by the following data:
\begin{itemize}[noitemsep,topsep=0pt]
\itemsep0em 
\item The vertex set $Z = \lbrace z_1, z_2, \hdots, z_n \rbrace$
\item A collection of simplices $\Sigma$, where $p$-dimensional simplex $\sigma_p$ is defined just as a $p+1$ element subset of~$Z$:
$$\sigma_p = \lbrace z_{i_1}, z_{i_2}, \hdots ,z_{i_{p+1}} \rbrace $$
\item We require that the collection $\Sigma$ is closed under taking \emph{faces}, that is for each $p$-dimensional simplex $\sigma_p$ all the $(p-1)$-dimensional simplices obtained by deleting one of the vertices $z_{i_1}, \hdots ,z_{i_p}$ are also elements of $\Sigma$.
\end{itemize}
\end{definition}
An example of a simplicial complex $\mathcal{S}$ is presented on \cref{fig:simp_compl}. It contains $5$ vertices $\lbrace z_1, z_2 \hdots ,z_5 \rbrace$ and several edges and faces: two-dimensional face $\sigma_2$ and one-dimensional edges $\sigma_1, \sigma_3, \sigma_4$. Note that these are \emph{maximal} simplices, since by the third property all the edges of $\sigma_2$ are also elements of $\mathcal{S}$. Important topological properties of $\mathcal{S}$ (such as connectedness, existence of one-dimensional loop) do not depend on in which Euclidean space $\mathcal{S}$ is embedded or on precise positions of vertices, but merely on the combinatorial data --- the number of points and which vertices together span a simplex. 

As was described in \cref{sec:idea} given a dataset $X$ sampled from a manifold $\mathcal{M}$ we would like to compute a family of simplicial complexes topologically approximating $\mathcal{M}$ on various scales, namely witness complexes. This family is defined as follows. First we choose some subset $L \subset X$ of points called landmarks (whereas points in $X$ are called witnesses) and some distance function $d(x, x')$, e.g., the ordinary Euclidean distance. There is not much theory about how to choose the best landmarks, but several strategies were proposed in \cite{de2004topological}. The first one is to choose landmarks sequentially by solving a certain minimax problem, and the second one is to just pick landmarks at random (by uniformly selecting a fixed number of points from $X$). We follow the second approach since the minimax strategy is known to have some flaws such as the tendency to pick up outliers. The selected landmarks will serve as the vertices of the simplicial complex and witnesses will help to decide on which simplices are inserted via a predicate ``is witnessed'':
\begin{equation} \label{eq:def-witness}
\begin{split}
\sigma \subset L  & \text{ is witnessed by } w\in X \text{ if } \forall l \in \sigma, \forall l' \in L \setminus \sigma \\ 
 & d(w, l)^2 \leq d(w, l')^2 + \alpha,  
\end{split}
\end{equation}
with $\alpha$ being a relaxation parameter which provides us with a sequence of simplicial complexes. The maximal value of $\alpha$ for the analysis is typically chosen to be proportional to the maximal pairwise distance between points in $L$. Witness complexes even for small values of $\alpha$ are good topological approximations to $\mathcal{M}$. The main advantage of a witness complex is that it allows constructing a reliable approximation using a relatively small number of simplices and makes the problem tractable even for large datasets. Even though it is known that in some cases it may fail to recover the correct topology \cite{boissonnat2009manifold}, it still can be used to compare topological properties of datasets, and if any better method is devised, we can easily replace the witness complex by this new more reliable simplicial complex.

\paragraph{Homology} 
The precise definition of the homology is technical, and we have to omit it due to the limited space. We refer the reader to [Chapter 2] \cite{hatcher2002algebraic} for a thorough discussion. 
The most important properties of homology can be summarized as follows. For any topological space $X$ the so-called $i^{th}$ homology groups $H_i$ are introduced. The actual number of $i$-dimensional holes in $X$ is given the \emph{rank} of $H_i$, the concept which is quite similar to the dimension of a vector space. These ranks are called the \emph{Betti numbers} $\beta_i$ and serve as a coarse numerical measure of homology. 

Homology is known to be one of the most easily computable topological invariants. In the case of $X$ being a simplicial complex $H_i$ can be computed by pretty much linear algebra, namely by analyzing kernels and images of certain linear maps.  Dimensions of matrices appearing in this task are equal to the numbers $d_k$ of simplices of specific dimension $k$ in $X$, e.g. in the case of \cref{fig:simp_compl} we have $d_0 = 5, d_1 = 6, d_2 = 1$ and matrices will be of sizes $6 \times 1$ and $5 \times 6$. Existent algorithms \cite{kaczynski2006computational} can handle extremely large simplicial complexes (with millions of simplices) and are available in numerous software packages. An important property of homology is that $k^{th}$ homology depends only on simplices of dimension at most $k+1$, which significantly speeds up computations.
\paragraph{Persistent homology}
In \cref{sec:idea} we discussed that to find a proxy of the correct topology of $\mathcal{M}$ it is insufficient to use single simplicial complex but rather a family of simplicial complexes is required. As we transition from one simplicial complex to another, some holes may appear, and some disappear. To distinguish between which are essential and which should be considered noise the concept of persistence was introduced \cite{edelsbrunner2000topological, zomorodian2005computing}. The formal Structure Theorem \cite{zomorodian2005computing} states that for each generator of homology (``hole'' in our notation) one could provide the time of its ``birth'' and ``death''. This data is pictorially represented as (\cref{fig:sc-family}, [bottom]), with the horizontal axis representing the parameter and the vertical axis representing various homology generators. To perform the computation of these barcodes, an efficient algorithm was proposed in \cite{zomorodian2005computing}. As an input to this algorithm one has to supply a sequence of $(\sigma_i, \varepsilon_i)$, with $\sigma_i$ being a simplex and $\varepsilon_i$ being its time of appearance in a family. This algorithm is implemented in several software packages such as \texttt{Dionysus} and \texttt{GUDHI}
\cite{maria2014gudhi}, but the witness complex is supported only in the latter.

\section{Algorithm \label{sec:stochastic}}
Let us now explain how we apply these concepts to construct a metric to compare the topological properties of two datasets. First let us define the key part of the algorithm -- the relative living times (RLT) of homology. Suppose that for a dataset $X$ and some choice of landmarks $L$ we have obtained a persistence barcode with the persistence parameter $\alpha$ spanning the range $[0, \alpha_{\max}]$. Let us fix the dimension $k$ in which we study the homology, and let $\mathcal{I}_k = \lbrace [b_i, d_i] \rbrace_{i=1}^n$ be the collection of persistence intervals in this dimension. Then in order to find the $k^{th}$ Betti number for a fixed value $\alpha$ one has to count the number of persistence intervals containing $\alpha$, and we obtain the integer valued function
\begin{equation}\label{eq:beta-def}
\beta_k(\alpha) \triangleq  \bigl|\lbrace [b_i, d_i] \in \mathcal{I}_k \colon \alpha 
\in [b_i, d_i] \rbrace \bigr|.
\end{equation}
Then the RLT are defined as follows (for non-negative integers $i$):
\begin{equation}\label{eq:rlt-def}
\RLT(i, k, X, L) \triangleq \frac{\mu \bigl(\lbrace \alpha \in [0, \alpha_{\max}] \colon \beta_k(\alpha) = i \rbrace \bigr)}{\alpha_{\max}},
\end{equation}
that it is for each possible value of $\beta_k(\alpha)$ we find how long it existed \emph{relatively} to the whole period of topological activity.
Note that in our analysis we use witness complexes which depend on the choice of landmarks, which is random. Thus it is reasonable to consider the distribution of $\RLT(i, k, X, L)$ on the set of landmarks (tuples of points), in other words, we repeatedly sample the landmarks and compute the RLT of the obtained persistence barcode. After sufficiently many experiments we can approximate the \emph{Mean} Relative Living Times (MRLT): 
\begin{equation}\label{eq:mrlt-def}
\mRLT(i, k, X) \triangleq \mathbb{E}_{L} [\RLT(i, k, X, L)].
\end{equation} 
We hypothesize that these quantities provide us with a good way to compare the topological properties of datasets, as they serve as measures of confidence in the estimation of the topology of the underlying manifolds. From \cref{eq:rlt-def} it follows that 
$$\sum_{i} \mRLT(i, k, X) = 1,$$
which suggest that for a fixed value of $k$ we could interpret $\mRLT(i, k, X)$ as a \emph{probability distribution} (over integers). This distribution defines our certainty about the number of $k$-dimensional holes in the underlying manifold of $X$ on \emph{average}.
In this work we consider the case $k=1$, i.e. we study the first homology of datasets. We motivate this by drawing an analogy with the Taylor series: we can get a good understanding of behavior of a function by looking at the first term of the series (see also \cite{ghrist2008barcodes} for discussion). Based on the probabilistic understanding given two datasets $X_1$ and $X_2$ we define a measure of their topological similarity (\emph{Geometry Score}) in the following way:
\begin{equation}\label{eq:topscore}
\begin{gathered}
\GeomScore(X_1, X_2) \triangleq \\ 
\sum_{i=0}^{i_{\max} - 1} \left(\mRLT(i, 1, X_1) - \mRLT(i, 1, X_2)\right)^2,
\end{gathered}
\end{equation}
with $i_{\max}$ being an upper bound on $\beta_1(\alpha)$ for $X_1$ and $X_2$ (for typical datasets we found that $i_{\max}=100$ suffices). 

To construct the witness complex given the sets of landmarks $L$ and witnesses $X$ one has to provide the matrix of pairwise distances between $L$ and $X$ and the maximal value of persistence parameter $\alpha$ (see \cref{eq:def-witness}). In our experiments, we have chosen $\alpha_{\max}$ to be proportional to the maximal pairwise distance between points in $L$ with some coefficient $\gamma$. Since we only compute $\beta_1(\alpha)$ the simplices of dimension at most $2$ are needed. In principle to compare two datasets any value of $\gamma$ suffices, however in our experiments we found that to get a reasonable distribution for datasets of size $\sim 5000$ the value $\frac{1}{128}$ yields good results (for large $\gamma$ a lot of time is spend in the regime of a single connected blob which shifts the distributions towards $0$).
We summarize our approach in \cref{alg:mrlt} and \cref{alg:top-score}. We also suggest that to obtain accurate results datasets of the same size should be used for comparison
\begin{algorithm}[htb!]
   \caption{The algorithm to compute RLT of a dataset. See \cref{sec:stochastic} for details. Suggested default values of the parameters for a dataset $X \in \mathbb{R}^{N \times D}$ are $L_0 = 64$, $\gamma = \frac{1}{128} / \frac{N}{5000}$, $i_{\max}=100$, $n=10000$.}
   \label{alg:mrlt}
\begin{algorithmic}
   \STATE {\bfseries Require:} $X$: $2D$ array representing the dataset
   \STATE {\bfseries Require:} $L_0$: Number of landmarks to use
   \STATE {\bfseries Require:} $\gamma$: Coefficient determining $\alpha_{\max}$
   \STATE {\bfseries Require:} $i_{\max}$: Upper bound on $i$ in $\RLT(i, 1, X, L)$
   \STATE {\bfseries Require:} $n$: Number of experiments
   \STATE {\bfseries Require:} \texttt{dist}$(A,\ B)$: Function computing the matrix of pairwise (Euclidean) distances between samples from $A$ and $B$
   \STATE {\bfseries Require:} \texttt{witness}$(d,\ \alpha,\ k)$: Function computing the family of witness complexes using the matrix of pairwise distances $d$, maximal value of persistence parameter $\alpha$ and maximal dimension of simplices $k$
   \STATE {\bfseries Require:} \texttt{persistence}$(w, k)$: Function computing the persistence intervals of a family $w$ in dimension $k$
   \STATE {\bfseries Returns:} An array of size $n \times i_{\max}$ of the obtained RLT for each experiment
   \STATE {\bfseries Initialize:} $\mathrm{rlt} = \texttt{zeros}(n,\ i_{\max})$
   \FOR{$i=0$ {\bfseries to} $n-1$}
   \STATE $L^{(i)} \leftarrow$ \texttt{random\_choice}$(X,\ \texttt{size}\negmedspace=\negmedspace L_0)$
   \STATE $d^{(i)} \leftarrow$  \texttt{dist}$(L^{(i)},\ X)$ 
   \STATE $\alpha^{(i)}_{\max} \leftarrow$ $\gamma \cdot  \texttt{max}(\texttt{dist}(L^{(i)},\ L^{(i)}))$
   \STATE $W^{(i)} \leftarrow \texttt{witness}(d^{(i)},\ \alpha^{(i)}_{\max},\ 2)$
   \STATE $\mathcal{I}^{(i)} \leftarrow \texttt{persistence}(W^{(i)},\ 1)$
   \FOR{$j=0$ {\bfseries to} $i_{\max} - 1$} 
   \STATE Compute $\RLT(j, 1, X, L^{(i)})$ using \cref{eq:rlt-def,eq:beta-def}
   \STATE $\mathrm{rlt}[i,\ j] \leftarrow \RLT(j, 1, X, L^{(i)})$
   \ENDFOR
   \ENDFOR
\end{algorithmic}
\end{algorithm}
\begin{algorithm}[htb!]
   \caption{\emph{Geometry Score}, the proposed algorithm to compute topological similarity between datasets}
   \label{alg:top-score}
\begin{algorithmic}
   \STATE {\bfseries Require:} $X_1, X_2$: arrays representing the datasets
   \STATE {\bfseries Returns:} $s$: a number representing the topological similarity of $X_1$ and $X_2$
   \STATE {\bfseries Initialize:} $s=0$
	\STATE For $X_1$ and $X_2$ run \cref{alg:mrlt} with the same collection of parameters, obtaining arrays $\mathrm{rlt}_1$ and $\mathrm{rlt}_2$
    \STATE $\mathrm{mrlt}_1 \leftarrow \texttt{mean}(\mathrm{rlt}_1,\ \texttt{axis}\negmedspace=\negmedspace 0)$
    \STATE $\mathrm{mrlt}_2 \leftarrow \texttt{mean}(\mathrm{rlt}_2,\ \texttt{axis}\negmedspace=\negmedspace 0)$
    \STATE $s \leftarrow \texttt{sum} ((\mathrm{mrlt}_1 - \mathrm{mrlt}_2)^2)$
\end{algorithmic}
\end{algorithm}
\paragraph{Complexity} Let us briefly discuss the complexity of each step in the main loop of \cref{alg:mrlt}. Suppose that we have a dataset $X \in \mathbb{R}^{N \times D}$. Computing the matrix of pairwise distances between all points in the dataset and the landmarks points requires $O(NDL_0)$ operations. The complexity of the next piece involving computing the persistence barcode is hard to estimate, however we can note that it does not depend on the dimensionality $D$ of the data. In practice this computation is done faster than computing the matrix in the previous step (for datasets of significant dimensionality). All the remaining pieces of the algorithm take negligible amount of time. This linear scaling of the complexity w.r.t dimensionality of the data allows us to apply our method even for high--dimensional datasets. On a typical laptop (3.1 GHz Intel Core i5 processor) one iteration of the inner loop of \cref{alg:mrlt} for one class of the MNIST dataset takes approximately $~900$ ms. 
\section{Experiments \label{sec:experiments}}
\begin{figure*}[h!]
\centering
\includegraphics[width=0.9\linewidth]{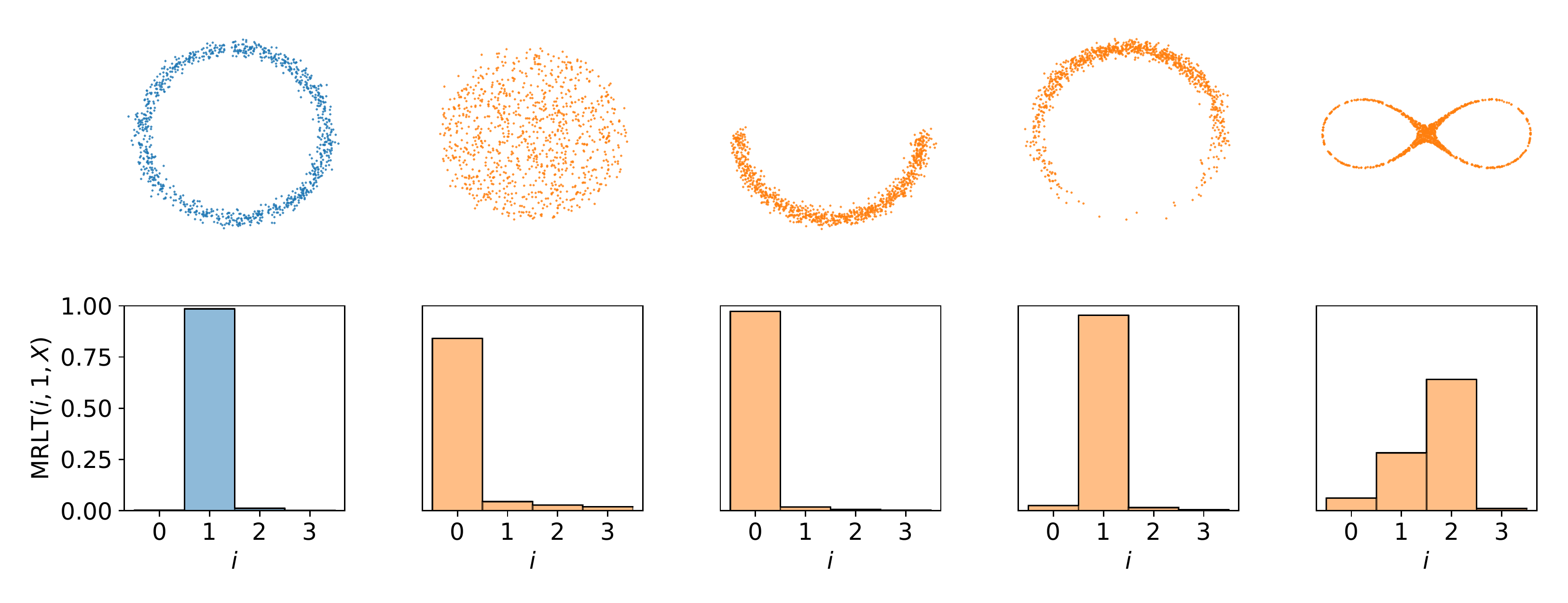}
 \caption{Mean Relative Living Times (MRLT) for various $2D$ datasets. The number of one-dimensional holes is correctly identified in all the cases. By comparing MRLT we find that the second dataset from the left is the most similar to the `ground truth' (noisy circle on the left).}
  \label{fig:synthetic}
\end{figure*}

\begin{figure*}[h!]
\centering
\includegraphics[width=1.0\linewidth]{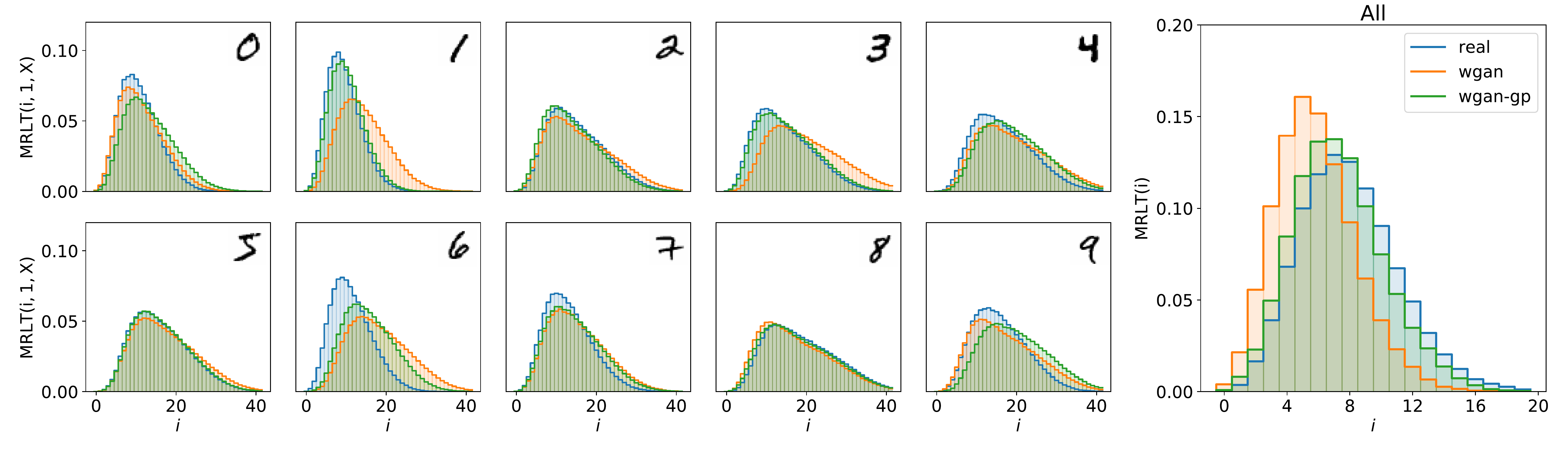}
 \caption{Comparison of MRLT of the MNIST dataset and of samples generated by WGAN and WGAN-GP trained on MNIST. MRLT match almost perfectly, however, WGAN-GP shows slightly better performance on most of the classes.} 
  \label{fig:mnist-mrlt}
\end{figure*}

\begin{figure}[htb!]
\centering
\includegraphics[width=1.0\linewidth]{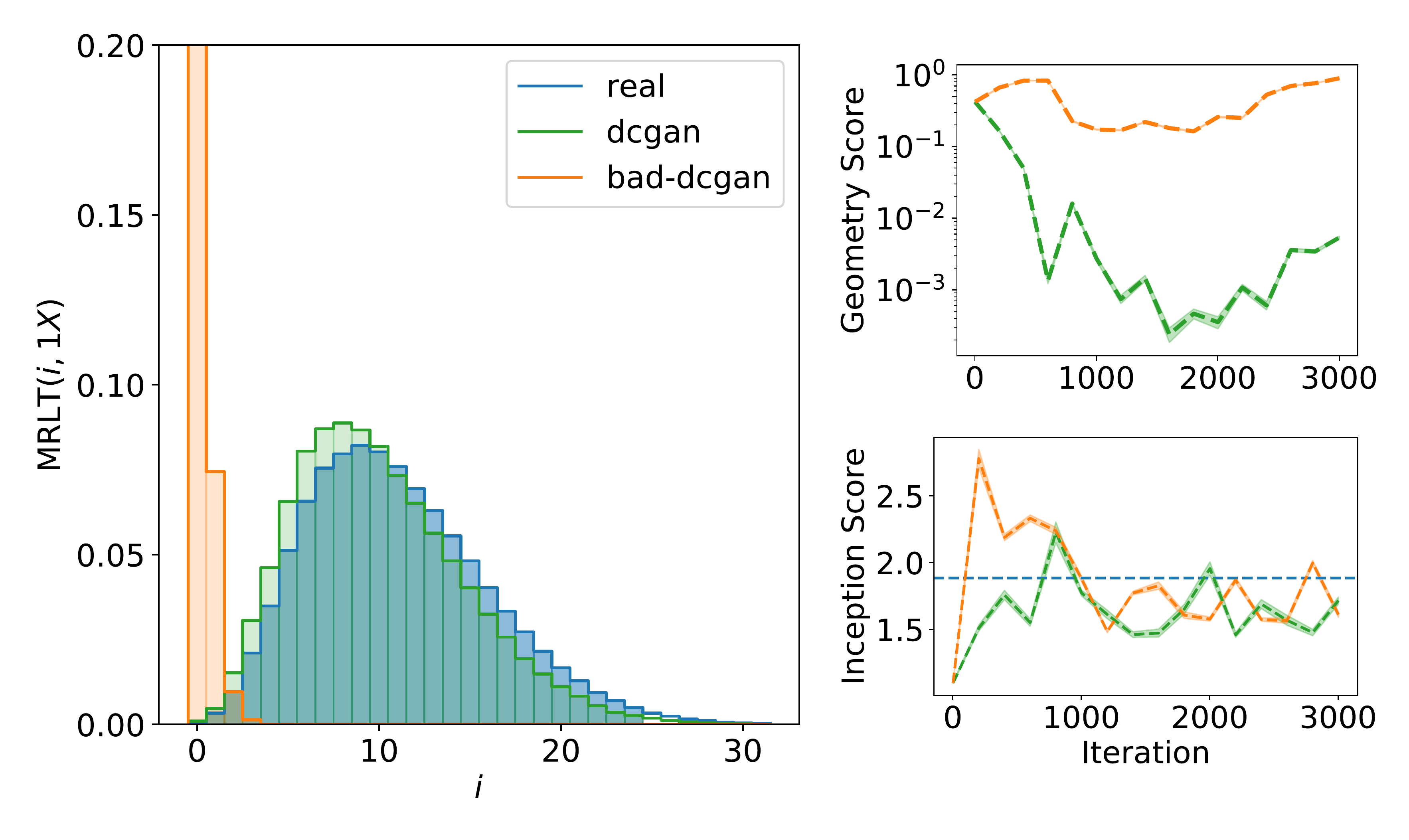}
 \caption{MRLT of the (cropped) CelebA dataset and samples generated using DCGAN and DCGAN with forced mode collapse. Plots on the right present the behavior of the \emph{Geometry Score} and \emph{Inception Score} for these two models during the training. Mode collapse in `bad-dcgan' is easily observable using the \emph{Geometry Score}.}
  \label{fig:celebA}
\end{figure}

\begin{figure}[h!]
\centering
\includegraphics[width=1.0\linewidth]{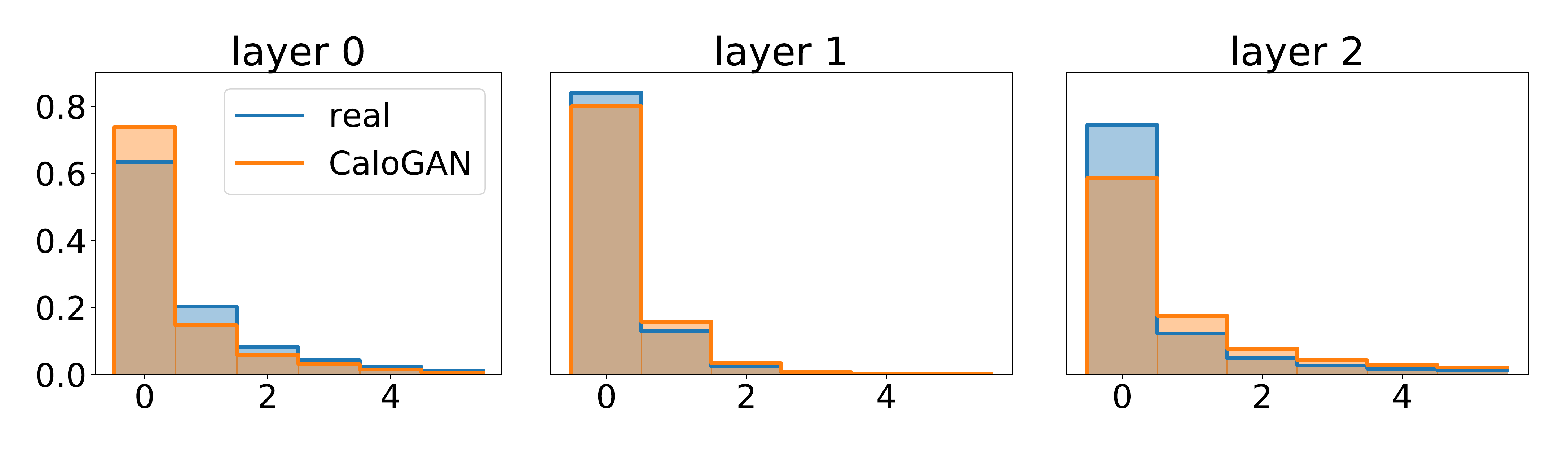}
 \caption{MRLT of the dataset used in experimental particle physics and of the samples generated using the corresponding CaloGAN model.} 
  \label{fig:calo-gan}
\end{figure}

\paragraph{Experimental setup}
We have implemented \cref{alg:mrlt,alg:top-score} in \texttt{Python} using \texttt{GUDHI}\footnote{\href{http://gudhi.gforge.inria.fr/}{http://gudhi.gforge.inria.fr/}} for computing witness complexes and persistence barcodes. Our code is available on \text{Github}\footnote{\href{https://github.com/KhrulkovV/geometry-score}{https://github.com/KhrulkovV/geometry-score}}. Default values of parameters in \cref{alg:mrlt} were used for experiments unless otherwise specified. We test our method on several datasets and GAN models:
\begin{itemize}
\item \textbf{Synthetic data} --- on synthetic datasets we demonstrate that our method allows for distinguishing the datasets based on their topological properties.   
\item \textbf{MNIST} --- as the next experiment we test our approach on the MNIST dataset of handwritten digits. We compare two recently proposed models: WGAN \cite{arjovsky2017wasserstein} and WGAN-GP \cite{gulrajani2017improved} in order to verify if the improved model WGAN-GP indeed produces better images.
\item \textbf{CelebA} --- to demonstrate that our method can be applied to datasets of large dimensionality we analyze the CelebA dataset \cite{liu2015faceattributes} and check if we can detect mode collapse in a GAN using MRLT.
\item \textbf{CaloGAN} --- as the final experiment we apply our algorithm to a dataset of a non-visual origin and evaluate the specific generative model CaloGAN \cite{paganini2017calogan}.
\end{itemize}
\paragraph{Synthetic data}
For this experiment we have generated a collection of simple $2D$ datasets $\lbrace X_j \rbrace_{j=1}^{5}$ (see \cref{fig:synthetic}) each containing $5000$ points.  As a test problem we would like to evaluate which of the datasets $\lbrace X_j \rbrace_{j=2}^{5}$ is the best approximation to the ground truth $X_1$. For each of $\lbrace X_j \rbrace_{j=1}^{5}$ we ran \cref{alg:mrlt} using $L_0=32, \ n=2000, \ i_{\max}=3, \ \gamma=\frac{1}{8}$ and compute MRLT using \cref{eq:mrlt-def}. The resulting distributions are visualized on \cref{fig:synthetic}, [bottom]. We observe that we can correctly identify the number of $1$-dimensional holes
in each space $X_j$ using the MAP estimate
\begin{equation}{\label{eq:correct-hom}}
\beta_1^{*}(X_j) = \underset{i}\argmax \mRLT(i, 1, X_j).
\end{equation}
It is clear that $X_4$ is the most similar dataset to $X_1$, which is supported by the fact that their MRLT are almost identical. Note that on such simple datasets we were able to recover the correct homology with almost $100 \text{\%}$ confidence and this will not be the case for more complicated manifolds in the next experiment.
\paragraph{MNIST} In this experiment we compare topological properties of the MNIST dataset and samples generated by the WGAN and WGAN-GP models trained on MNIST.  It was claimed that the WGAN-GP model produces better images and we would like to verify if we can detect it using topology.
For the GAN implementations we used the code\footnote{\href{https://github.com/igul222/improved_wgan_training}{https://github.com/igul222/improved\_wgan\_training}} provided by the authors of \cite{gulrajani2017improved}. We have trained each model for $25$ epochs and generated $60000$ samples. To compare topology of each class individually we trained a CNN classifier on MNIST (with $~99.5 \text{\%}$ test accuracy) and split generated datasest into classes (containing roughly $6000$ images each). For every class and each of the $3$ corresponding datasets (`base', `wgan', `wgan--gp') we run \cref{alg:mrlt} and compute MRLT with $\gamma = \frac{1}{128}$. Similarly we evaluate MRLT for the entire datasets without splitting them into classes using $\gamma = \frac{1}{1000}$. The obtained MRLT are presented on \cref{fig:mnist-mrlt} and the corresponding 
\emph{Geometry Scores} for each model are given in \cref{tab:mnist-table}. We observe that both models produce distributions which are very close to the ground truth, but for almost all classes WGAN-GP shows better scores. We can also note that for the entire datasets (\cref{fig:mnist-mrlt}, [right]) the predicted values of homology does not seem to be much bigger than for each individual digit. One possible explanation is that some samples (like say of class `$7$') fill the holes in the underlying manifolds of other classes (like class `$1$' in this case) since they look quite similar. 
\setlength{\tabcolsep}{5.pt}
\begin{table*}[h!]
\caption{\emph{Geometry Scores} $\times 10^3$ of WGAN and WGAN-GP trained on the MNIST dataset (see also \cref{fig:mnist-mrlt}). Each class contained roughly $6000$ images, except for `All' which corresponds to the total datasets of $60000$ images.}
\label{tab:mnist-table}
\vskip 0.15in
\begin{center}
\begin{small}
\begin{sc}
\begin{tabular}{lccccccccccc}
\toprule
Label & 0 & 1 & 2 & 3 & 4 & 5 & 6 & 7 & 8 & 9  & all \\
\midrule
WGAN    & \textbf{0.85}&21.4&0.60&7.04&\textbf{1.52}&0.47&22.8&2.20&0.76&\textbf{1.27}&26.1 \\
WGAN-GP & 5.19&\textbf{1.44}&\textbf{0.54}&\textbf{0.27}&2.16&\textbf{0.03}&\textbf{13.5}&\textbf{1.38}&\textbf{0.14}&5.00&\textbf{2.04} \\
\bottomrule
\end{tabular}
\end{sc}
\end{small}
\end{center}
\vskip -0.1in
\end{table*}
\paragraph{CelebA} We now analyze the popular CelebA dataset consisting of photos of various celebrities. In this experiment we would like to study if we can detect mode collapse using our method. To achieve this we train two GAN models --- a good model with the generator having high capacity and a second model with the generator much weaker than the discriminator.
In this experiment we utilize the DCGAN model and use the implementation provided\footnote{\href{https://github.com/carpedm20/DCGAN-tensorflow}{https://github.com/carpedm20/DCGAN-tensorflow}} by the authors \cite{radford2015unsupervised}. For the first model (`dcgan') we use the default settings, and for the second (`bad-dcgan') we set the latent dimension to $8$ and reduce the size of the fully connected layer in the generator to $128$ and number of filters in convolutional layers to $4$. Images in the dataset are of size $108 \times 108$ and to obtain faces we perform the central crop which reduces the size to $64 \times 64$. We trained both models for $25$ epochs and produced $10000$ images for our analysis. Similarly, we randomly picked $10000$ (cropped) images from the original dataset. We report the obtained results on \cref{fig:celebA}. MRLT obtained using the good model matches the ground truth almost perfectly and \emph{Geometry Score} of the generated dataset is equal to $14 \times 10^{-3}$, confirming the good visual quality of the samples \cite{radford2015unsupervised}. MRLT obtained using the weak model are maximized for $i=0$, which suggests that the samples are either identical or present very little topological diversity (compare with \cref{fig:plane-topology}), which we confirmed visually. On \cref{fig:celebA}, [right] we report the behavior of the \emph{Geometry Score} and \emph{Inception Score} \cite{improved_techniques} w.r.t the iteration number. The \emph{Inception Score} introduced uses the pretrained Inception network \cite{inception} and is defined as 
\begin{equation*}\label{eq:incep-score}
I(\lbrace x_n \rbrace_{n=1}^{N}) \triangleq \exp \mathbb{E}_{\textbf{x}}(\mathrm{D_{KL}}(p(y | \textbf{x}) || p(y))),
\end{equation*}
where $p(y | \textbf{x})$ is approximated by the Inception network and $p(y)$ is computed as $p(y) = \frac{1}{N} \sum_i p(y | \textbf{x}_i)$. Note that the \emph{Geometry Score} of the better model rapidly decreases and of the mode collapsed model stagnates at high values. Such behavior could not be observed in the \emph{Inception Score}. 
\vspace{-2.5mm}
\paragraph{CaloGAN} In this experiment, we will apply our technique to the dataset appearing in the experimental particle physics. This dataset\footnote{\label{note1}\href{https://data.mendeley.com/datasets/pvn3xc3wy5/1}{https://data.mendeley.com/datasets/pvn3xc3wy5/1}} represents a collection of a calorimeter (an experimental apparatus measuring the energy of particles) responses, and it was used to create a generative model \cite{paganini2017calogan} in order to help physicists working at the LHC. Evaluating the obtained model\footnote{\label{note2}\href{https://github.com/hep-lbdl/CaloGAN}{https://github.com/hep-lbdl/CaloGAN}} is a non-trivial task and was performed by comparing physical properties of the obtained and the real data. Since our method is not limited to visual datasets we can apply it in order to confirm the quality of this model. For the analysis we used `eplus' dataset which is split into $3$ parts (`layer 0', `layer 1', `layer 2') containing matrices of sizes $3 \times 96, 12 \times 12, 12 \times 6$ correspondingly. We train the CaloGAN model with default settings for $50$ epochs and generate $10000$ samples (each sample combines data for all $3$ layers). We then randomly pick $10000$ samples from the original dataset and compare MRLT of the data and generated samples for each layer. Results are presented on \cref{fig:calo-gan}. It appears that topological properties of this dataset are rather trivial, however, they are correctly identified by CaloGAN. Slight dissimilarities between the distributions could be connected to the fact that the physical properties of the generated samples do not exactly match those of the real ones, as was analyzed by the authors of \cite{paganini2017calogan}.
\vspace{-2.5mm}
\section{Related work and discussion\label{sec:related}}
Several performance measures have been introduced to assess the performance of GANs used for natural images. \emph{Inception Score} \cite{improved_techniques} uses the outputs of the pretrained Inception network, and a modification called \emph{Fr\'echet Inception Distance (FID) }\cite{heusel2017gans} also takes into account second order information of the final layer of this model. Contrary to these methods, our approach does not use auxiliary networks and is not limited to visual data. We note, however, that since we only take topological properties into account (which do not change if we say shift the entire dataset by $1$) assessing the \emph{visual} quality of samples may be difficult based only on our algorithm, thus in the case of natural images  we propose to use our method in conjunction with other metrics such as \emph{FID}. We also hypothesize that in the case of the large dimensionality of data \emph{Geometry Score} of the features extracted using some network will adequately assess the performance of a GAN.
\section{Conclusion}
We have introduced a new algorithm for evaluating a generative model. We show that the topology of the underlying manifold of generated samples may be different from the topology of the original data manifold, which provides insight into properties of GANs and can be used for hyperparameter tuning. We do not claim however that the obtained metric correlates with the visual quality as estimated by humans and leave the analysis to future work. We hope that our research will be useful to further theoretical understanding of GANs.
\newpage
\section*{Acknowledgements}
We would like to thank the anonymous reviewers for their valuable comments. We also thank Maxim Rakhuba for productive discussions and making our illustrations better. This study was supported by the Ministry of Education and Science of the Russian Federation (grant
14.756.31.0001).
\bibliography{main}
\bibliographystyle{icml2018}
\end{document}